\documentclass[conference]{IEEEtran}
\IEEEoverridecommandlockouts
\usepackage{cite}
\usepackage{amsmath,amssymb,amsfonts}
\usepackage{algorithmic}
\usepackage{graphicx}
\usepackage{textcomp}
\usepackage{multirow}
\usepackage{colortbl,xcolor}
\usepackage{booktabs}
\usepackage[table,xcdraw]{xcolor}

\usepackage{hyperref}
\usepackage{fontawesome5}

\def\BibTeX{{\rm B\kern-.05em{\sc i\kern-.025em b}\kern-.08em
    T\kern-.1667em\lower.7ex\hbox{E}\kern-.125emX}}

\begin{document}
\title{GeoMag: Geometric-Aware Video Motion Magnification via State Space Model}
\graphicspath{{images/}}

\author{
\IEEEauthorblockN{
Kecheng Han$^{1,\dagger}$ \quad
Yuchen Zhang$^{1,\dagger}$ \quad
Bingqing Liu$^{1}$ \quad
Boqiang Guo$^{1}$ \quad 
Wenbin Zheng$^{1,*}$ \quad
Shiyuan Pei$^{2,*}$
}
\IEEEauthorblockA{
$^{1}$School of Software Engineering, Xi'an Jiaotong University \\
$^{2}$Xi'an Jiaotong University \\
xjhkc@stu.xjtu.edu.cn \quad yczhang@stu.xjtu.edu.cn \quad wenbin@xjtu.edu.cn \\
\small $^\dagger$Kecheng Han and Yuchen Zhang contributed equally to this work. \\
\small $^*$Wenbin Zheng and Shiyuan Pei are the corresponding authors. \\
\\
\faGithub\ \href{https://github.com/scottHankcheng/GeoMag}{\texttt{github.com/scottHankcheng/GeoMag}}
}
}
\maketitle

\begin{abstract}
Video Motion Magnification (VMM) reveals imperceptible dynamics but often suffers from structural inconsistencies under complex geometric transformations. Existing learning-based methods generally face a trade-off between the limited global context of CNNs and the high computational cost of Transformers. In addition, current training protocols, largely dominated by simple linear motion, fail to capture the geometric and imaging complexities encountered in real-world videos. To address these issues, we propose GeoMag, a geometric-aware VMM framework built upon State Space Models to achieve globally consistent motion amplification with linear complexity. We further construct Geo-200K, a large-scale synthetic dataset that introduces rich geometric transformations together with sensor-realistic degradations, improving the diversity and realism of training signals. Extensive experiments on synthetic and real-world benchmarks show that GeoMag consistently outperforms prior methods in visual fidelity and computational efficiency, while producing fewer artifacts and better structural consistency. The source code is available at https://github.com/scottHankcheng/GeoMag
\end{abstract}

\vspace{-0.15cm}

\section{Introduction}

Subtle motions encode critical physical information yet remain imperceptible at sub-pixel scales~\cite{le2019seeing}. Video Motion Magnification (VMM) serves as a ``motion microscope''~\cite{liu2005motion}, widely used in structural~\cite{zhang2023hybrid}, biomedical~\cite{shabi2020motion}, and micro-expression analysis~\cite{bai2021micro,Xiaoicmemicro,fang2023rmesICME,icmeGraph,icmeMicro}. However, these minute dynamics are often commensurate with imaging noise~\cite{zhou2022audio, oh2018learning}, leading to a core challenge: indiscriminately amplifying motion often boosts noise, causing severe visual artifacts. Reliable VMM requires separating meaningful dynamics from complex backgrounds without compromising structural integrity~\cite{2505.17476}.

Traditional Eulerian approaches utilize spatial decomposition with hand-crafted filters~\cite{wu2012eulerianfluid}. While subsequent phase-based methods~\cite{wadhwa2013phase,wadhwa2014riesz,2018jerk} improved signal isolation, they often suffer from ringing artifacts and require manual frequency specification, limiting their applicability to time-varying dynamics~\cite{wadhwa2013phase}. Consequently, the field has pivoted towards data-driven deep learning paradigms.

Learning-based paradigms~\cite{oh2018learning,2603.01993} have significantly improved reconstruction quality over traditional methods, but achieving robust VMM remains constrained by a structural dilemma. Specifically, the task demands global receptive fields to maintain geometric consistency, yet simultaneously requires selective sensitivity to avoid amplifying high-frequency imaging noise. Existing architectures struggle to reconcile these conflicting needs. CNNs~\cite{oh2018learning,LDSSVMM2025,singh2023multi}, while efficient, lack the long-range dependency modeling required for geometric coherence, often leading to local distortions. Conversely, Transformers~\cite{wang2024eulermormer,STBVMM_LADOROIGE2023110493,wang2024frequency} offer global context but suffer from quadratic computational complexity. Moreover, although Transformers provide global context, their attention mechanisms are not explicitly designed to distinguish subtle motion cues from irrelevant sensor noise, which may limit their robustness in VMM.

Beyond architecture, training data quality represents another critical bottleneck. Most methods rely on the LBVMM protocol~\cite{oh2018learning}, where synthetic motions are largely restricted to simple linear translation. This limitation fails to capture the geometric and imaging complexities encountered in practice, including rotation, compound motion, and sensor degradation, which may cause models to overfit to overly simplified motion patterns. Consequently, establishing a high-quality dataset characterized by richer geometric diversity is important for achieving superior magnification fidelity.

In this paper, we present a unified solution that combines data synthesis with architectural design. We introduce Geo-200K, a large-scale dataset constructed from CocoNut~\cite{deng2024coconut} and OpenImages~\cite{openiamges2020open} that explicitly incorporates complex geometric transformations together with sensor-realistic degradations. Complementing this, we propose GeoMag to leverage State Space Models for robust amplification. Our main contributions are:

\begin{itemize}
    \item We propose GeoMag, a novel framework based on State Space Models designed for subtle motion magnification. Our approach effectively handles complex geometric motions while reducing visual artifacts and blurring compared to existing methods.
     \item We construct Geo-200K, a large-scale synthetic dataset incorporating complex geometric transformations and sensor-realistic degradations, providing more diverse and realistic supervision for VMM training.
    \item Extensive experiments validate the efficacy of the proposed framework and the established dataset, demonstrating that GeoMag surpasses state-of-the-art methods in visual fidelity and efficiency, while Geo-200K provides more realistic and challenging supervision for training.

\end{itemize}

\section{Related Work}

\noindent \textbf{Hand-crafted Magnification Filters.}
Early research originated from the Lagrangian perspective~\cite{liu2005motion} but proved computationally expensive. Consequently, the field shifted to the Eulerian perspective~\cite{wu2012eulerianfluid,wadhwa2013phase}, focusing on pixel variations within fixed regions using Laplacian or steerable pyramids. While pioneering, these methods rely heavily on hand-crafted priors and manual frequency specification. This reliance limits their robustness against noise and occlusions, often necessitating extensive, scene-specific hyperparameter tuning that hinders practical deployment~\cite{oh2018learning,2018jerk}.

\noindent \textbf{Deep Learning for VMM.} Data-driven approaches have transcended the limitations of hand-crafted priors by learning disentangled representations. Pioneering work by Oh et al.~\cite{oh2018learning} formulated VMM as a spatial decomposition task using convolutional networks. 
Subsequent studies further refined this paradigm through multi-scale architectures~\cite{singh2023multi}, or specialized axial-based strategies such as axial magnification~\cite{2024Axial} and learnable axial filters~\cite{LDSSVMM2025}. However, restricted by local receptive fields, these methods lack global context, often confusing subtle texture variations with motion and causing spatial artifacts.
To capture long-range dependencies, recent works~\cite{wang2024eulermormer,STBVMM_LADOROIGE2023110493} utilize self-attention mechanisms. However, the quadratic complexity imposes prohibitive memory costs, rendering high-resolution processing—essential for resolving minute motion—computationally intractable.

\noindent \textbf{State Space Models in Vision.} 
Architectures like Mamba~\cite{gu2024mamba} achieve global receptive fields with linear complexity via selective scanning. 
This paradigm has been rapidly adapted into efficient backbones for high-level understanding tasks~\cite{zhu2024visionmambaefficientvisual}, and further extended to low-level image restoration~\cite{guo2024mambairsimplebaselineimage}.
Leveraging this, we construct a dedicated VMM architecture that captures the global context essential for precise signal separation, effectively avoiding the computational bottlenecks of attention mechanisms.
\section{Synthetic Training Dataset}

\begin{figure}[htbb]
  \centering
    \vspace{-0.15cm}
  \includegraphics[width=\linewidth]{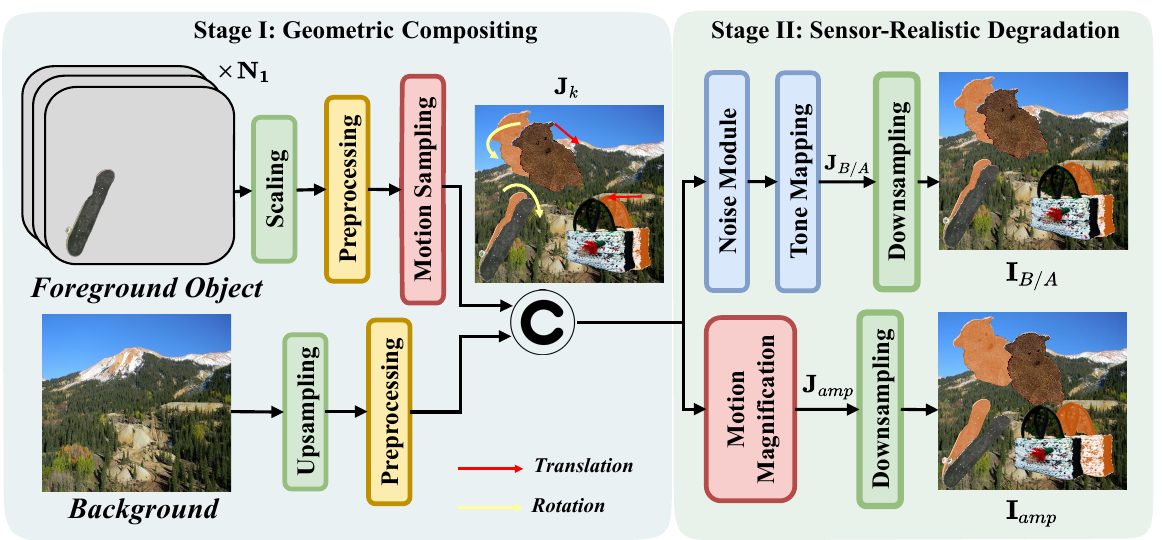}
  \vspace{-0.5cm}
 \caption{Overview of the synthetic pipeline operating in the linear HR domain. Stage I composites foregrounds with backgrounds, simulating diverse motions: rotation (skateboard), translation (bag), and combined (bear) as shown in $\mathbf{J}_k$. Stage II generates realistic noisy inputs $\mathbf{I}_{B/A}$ and clean magnified ground truth $\mathbf{I}_{amp}$. Orange silhouettes indicate original object positions.}
\label{fig:datasetOverview}
 \vspace{-0.25cm}
\end{figure}

Obtaining paired real-world data of subtle motions and their magnified counterparts is practically intractable. Therefore, we meticulously establish a comprehensive physics-based synthesis pipeline (Fig.~\ref{fig:datasetOverview}) to generate high-quality training pairs. 

To ensure textural realism and semantic diversity, real-image assets are employed. Specifically, for each generated scene, $N_1$ segmented foreground objects (where $N_1$ is randomly selected between 8 and 12) are randomly drawn from CocoNut~\cite{deng2024coconut} along with background images (denoted as $\mathbf{B}$) from OpenImages~\cite{openiamges2020open}. These assets serve as the fundamental components for the proposed pipeline, which is orchestrated into two sequential stages, commencing with geometric compositing.

\noindent \textbf{Stage I: High-Fidelity Geometric Compositing.}
Stage~I operates entirely in the linear high-resolution (HR) domain to strictly obey photometric linearity. A background image is converted from sRGB to linear space, normalized to $[0,1]$, and upsampled to $(sH)\times(sW)$ using Lanczos interpolation to yield $\mathbf{B}$. To bridge the domain gap, we apply mild Gaussian blurring and noise injection to $\mathbf{B}$, statistically matching it to the foreground elements.

Foreground objects undergo Adaptive Scaling and Statistical Alignment to align their brightness and sharpness with the background, resulting in the object set $\{\mathbf{O}_i\}_{i=1}^{N_1}$ with corresponding soft masks $\{\mathbf{M}_i\}$. For motion generation, each object $\mathbf{O}_i$ is assigned a rigid transformation $\mathcal{T}_{i,k} = \{R_{i,k}, \mathbf{t}_{i,k}\}$ for frame $k \in \{A,B\}$. To ensure a diverse range of dynamics, we adopt a specific sampling ratio for motion types: 30\% for pure translation, 30\% for pure rotation, and 40\% for combined transformation. Additionally, to ensure physical plausibility, we impose strict dual-stage kinematic constraints on the input displacement $\Delta \mathbf{p}_{in}$ and rotation $\theta_{in}$, as well as their magnified states (scaled by the magnification factor $\alpha$):

\begin{equation}
\begin{aligned}
    & \|\Delta \mathbf{p}_{in}\|_2 \le 3\,\text{px}, \quad & |\theta_{in}| \le 5^\circ; \\
    & \|\alpha \cdot \Delta \mathbf{p}_{in}\|_2 \le 30\,\text{px}, \quad & |\alpha \cdot \theta_{in}| \le 10^\circ.
\end{aligned}
\end{equation}
These bounds prevent numerical instabilities. Let $\mathcal{C}(\cdot)$ denote the sub-pixel affine compositing operator. The HR composite $\mathbf{J}_k$ (as labeled in Fig.~\ref{fig:datasetOverview}) is formulated as:
\begin{equation}
    \mathbf{J}_k = \mathcal{C}\Big( \mathbf{B}, \big\{ (\mathbf{O}_i, \mathbf{M}_i, \mathcal{T}_{i,k}) \big\}_{i=1}^{N_1} \Big).
\end{equation}

\begin{figure*}[!tbp]
  \centering
  \includegraphics[width=0.9\linewidth]{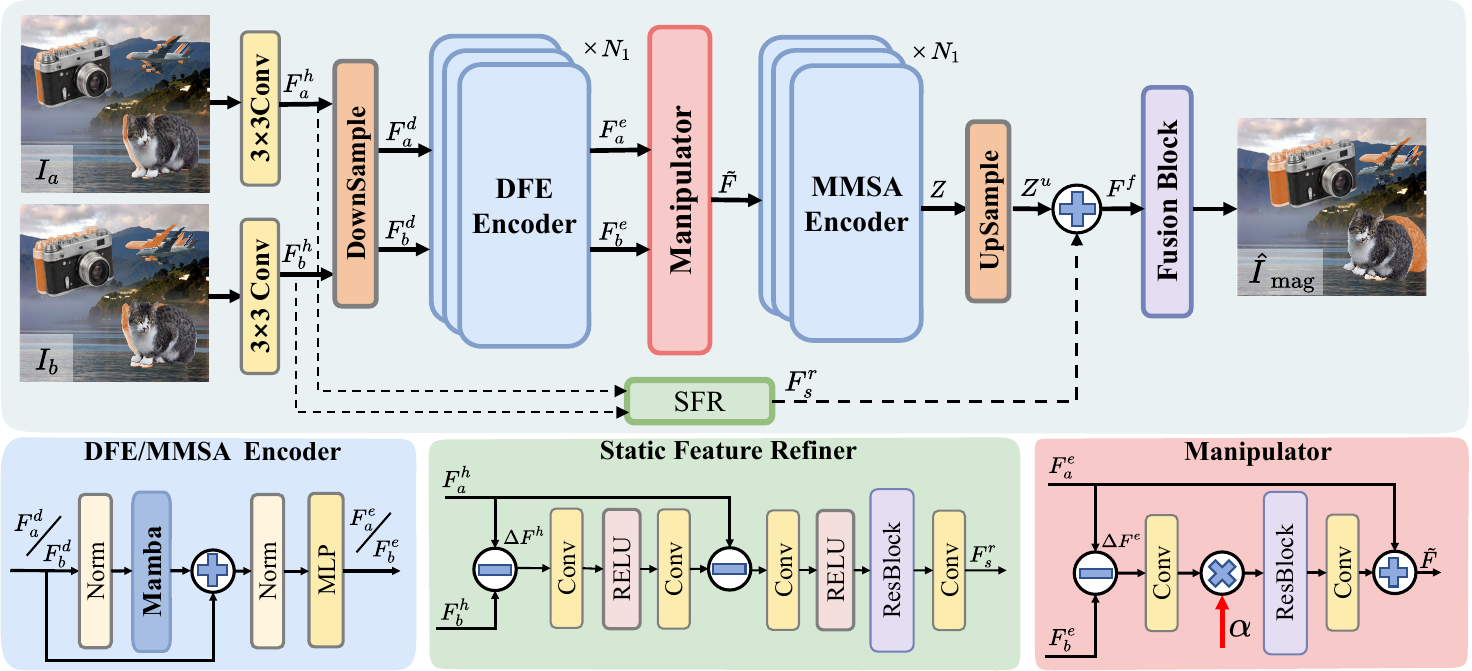}
  \vspace{-0.25cm}
 \caption{The overall architecture of the proposed GeoMag framework. Given a source frame $I_a$ and a subsequent frame $I_b$, the network synthesizes a magnified frame $\hat{I}_a$ where the motion is amplified by a factor $\alpha$. The pipeline consists of a dual-stream design: (1) a primary motion stream utilizing the Deep Feature Extractor (DFE) and Mamba-based Sequence Awareness (MMSA) encoder to capture globally consistent motion dynamics via State Space Models; and (2) a parallel Static Feature Refiner (SFR) stream designed to preserve high-frequency details. The Manipulator interconnects the encoders to execute motion amplification. The orange semi-transparent overlays highlight the original object positions.}
 \vspace{-0.45cm}
\label{fig:model_arch}
\end{figure*}

\noindent \textbf{Stage II: Sensor-Realistic Degradation.}
Stage~II maps the HR composites to the observed low-resolution (LR) space. 
The input composites $\mathbf{J}_k$ first undergo a Noise Module (simulating photon shot noise and readout artifacts), followed by Tone Mapping $\Gamma(\cdot)$ to convert the signal to the sRGB domain. 
As shown in the upper branch of Fig.~\ref{fig:datasetOverview}, this yields the degraded high-resolution intermediates $\mathbf{J}_{B/A}$. 
Subsequently, these frames undergo optical anti-aliasing and Downsampling $\Phi_{down}$.
To simulate the Analog-to-Digital Converter (ADC), we apply a quantization function $\mathcal{Q}(\cdot)$. The final input pair $\mathbf{I}_{B/A}=\{\mathbf{I}_A, \mathbf{I}_B\}$ is obtained via:
\begin{equation}
    \mathbf{I}_k = \mathcal{Q}\Big( \Phi_{down}\big( \Gamma(\mathbf{J}_k + \mathbf{N}_{sensor}) \big) \Big).
\end{equation}

In the supervision branch, we generate the ground truth via a parallel noise-free path. We construct an explicitly magnified HR frame, labeled as $\mathbf{J}_{amp}$, using the amplified motion $\mathcal{T}^\alpha$. This frame undergoes the same optical downsampling but excludes sensor noise. The final magnified ground truth $\mathbf{I}_{amp}$ is formulated as:
\begin{equation}
  \mathbf{I}_{amp} = \mathcal{Q}\big( \Phi_{down}( \Gamma(\mathbf{J}_{amp}) ) \big).
\end{equation}
This pipeline ensures that our model is trained on data reflecting both complex geometric kinematics and realistic imaging physics.

\section{Method}
\subsection{Overview}
\begin{table*}[!htbp]
\centering
\renewcommand{\arraystretch}{1.2} 
\caption{Quantitative comparison on four synthetic datasets. Group I reports official baselines, Group II reports retrained baselines on Geo-200K, and Group III reports our framework.}
\vspace{-0.25cm}
\resizebox{\textwidth}{!}{%
\begin{tabular}{c|c|ccc|ccc|ccc|ccc}
\hline
\multirow{2}{*}{\textbf{Method}} & \multirow{2}{*}{\textbf{Venue}} & \multicolumn{3}{c|}{\textbf{Synthetic-I} (Translation)} & \multicolumn{3}{c|}{\textbf{Synthetic-II} (Rotation)} & \multicolumn{3}{c|}{\textbf{Synthetic-III} (Noise)} & \multicolumn{3}{c}{\textbf{Synthetic-IV} (Blur)} \\ 
& & RMSE $\downarrow$ & PSNR $\uparrow$ & LPIPS $\downarrow$ & RMSE $\downarrow$ & PSNR $\uparrow$ & LPIPS $\downarrow$ & RMSE $\downarrow$ & PSNR $\uparrow$ & LPIPS $\downarrow$ & RMSE $\downarrow$ & PSNR $\uparrow$ & LPIPS $\downarrow$ \\ \hline

\multicolumn{14}{l}{\textit{\textbf{Group I: Existing Baselines (Official implementations)}}} \\ \hline
Phase~\cite{wadhwa2013phase} & TOG’13 & 46.36 & 15.29 & 0.48 & 45.28 & 15.53 & 0.46 & 45.36 & 15.50 & 0.49 & 45.65 & 15.46 & 0.49 \\ 
Jerk~\cite{2018jerk} & CVPR’18 & 45.34 & 15.84 & 0.42 & 44.32 & 16.38 & 0.44 & 44.20 & 15.58 & 0.46 & 45.69 & 15.50 & 0.42 \\
LBVMM~\cite{oh2018learning} & ECCV’18 & 35.59 & 16.16 & 0.34 & 33.76 & 16.78 & 0.35 & 34.55 & 18.82 & 0.38 & 34.70 & 18.76 & 0.38 \\ 
FlowMag~\cite{flowmag_2023} & NIPS’2023 & 34.83 & 17.68 & 0.32 & 32.02 & 18.54 & 0.28 & 32.45 & 18.40 & 0.30 & 32.58 & 18.38 & 0.30 \\ 
STB-VMM~\cite{STBVMM_LADOROIGE2023110493} & KBS’2023 & 28.92 & 18.95 & 0.35 & 29.40 & 20.25 & 0.33 & 26.43 & 20.25 & 0.33 & 26.60 & 20.21 & 0.34 \\ 
EulerMormer~\cite{wang2024eulermormer} & AAAI’2024 & 28.70 & 19.48 & 0.29 & 26.32 & 20.40 & 0.26 & 27.59 & 19.94 & 0.28 & 27.72 & 19.91 & 0.28 \\ 
LBAVMM~\cite{2024Axial} & ECCV’2024 & 30.62 & 18.79 & 0.40 & 28.30 & 19.60 & 0.38 & 28.34 & 19.59 & 0.39 & 28.46 & 19.55 & 0.39 \\
LDSSVMM
~\cite{LDSSVMM2025} & KBS’2025 & 32.80 & 17.89 & 0.31 &31.82 & 18.26 & 0.29 & 31.69 & 18.29 & 0.30 & 31.62 & 18.30 & 0.30 \\\hline

\multicolumn{14}{l}{\textit{\textbf{Group II: Impact of Geo-200K Data (Retrained Baselines)}}} \\ \hline
LBVMM (Retrained) & - & 34.30 & 16.96 & 0.32 & 32.63 & 17.18 & 0.34 & 32.35 & 19.68 & 0.36 & 34.70 & 19.32 & 0.35 \\ 
STB-VMM (Retrained) & - & \underline{27.15} & \underline{19.63} & \underline{0.28} & 28.62 & \underline{20.78} & \underline{0.26} & \underline{25.65} & \underline{20.76} & \underline{0.25} & \underline{25.96} & \underline{20.96} & \underline{0.26} \\ \hline

\multicolumn{14}{l}{\textit{\textbf{Group III: Proposed Framework}}} \\ \hline
Ours (Original Data) & - & 28.07 & 19.62 & 0.19 & \underline{25.90} & 20.44 & 0.18 & 26.12 & 20.34 & 0.18 & 26.27 & 20.30 & 0.19 \\ 
\textbf{Ours (Geo-200K)} & - & \textbf{25.95} & \textbf{21.32} & \textbf{0.18} & \textbf{23.61} & \textbf{21.35} & \textbf{0.16} & \textbf{23.45} & \textbf{21.39} & \textbf{0.17} & \textbf{24.91} & \textbf{21.83} & \textbf{0.18} \\ \hline

\end{tabular}%
}
\label{tab:MainCompTable}
\vspace{-0.4cm}
\end{table*}

\noindent \textbf{Overall Framework.} Given a source frame $I_a$ and a driving frame $I_b$, along with a magnification factor $\alpha$, the primary objective is to synthesize a high-fidelity magnified frame $\hat{I}_a$. The framework is defined as a mapping function that amplifies the geometric motion from $I_b$ while preserving the static details of $I_a$. As illustrated in Figure \ref{fig:model_arch}, the GeoMag architecture adopts a dual-stream design: a primary stream for global motion dynamics (comprising DFE, Manipulator, and MMSA) and a secondary stream for static refinement (SFR).
\vspace{-0.3cm}

\subsection{Feature Disentanglement and Manipulation}
\noindent \textbf{Deep Feature Encoding.} To isolate motion signals from appearance, the input frames are first projected into a latent space. Shallow features $F_a^h$ and $F_b^h$ are extracted and downsampled to $F_a^d$ and $F_b^d$. The DFE Encoder then processes these maps to derive deep semantic representations, denoted as $F_a^e$ and $F_b^e$. Specifically, the DFE is constructed by stacking $N_1$ Vision Mamba blocks (where $N_1=12$), allowing it to capture the complex structural geometry necessary for robust motion analysis.

\noindent \textbf{Motion Manipulation.} The core amplification occurs within the Manipulator module. Distinct from traditional frequency-domain phase manipulation, the proposed module computes motion differentials directly within the latent feature space. This feature differential is linearly scaled by the magnification factor $\alpha$ and subsequently processed by a non-linear manipulation function $\mathcal{M}$, which denotes the convolution and residual refinement operations implemented in the Manipulator block shown in Fig.~\ref{fig:model_arch}. The magnified feature $\tilde{F}$ is obtained by adding this amplified motion back to the source representation:
\begin{equation}
    \tilde{F} = F_a^e + \mathcal{M}\left( \alpha \cdot (F_b^e - F_a^e) \right)
\end{equation}
While this compact formulation focuses manipulation on dynamic components, the raw differential $(F_b^e - F_a^e)$ inherently contains both geometric shifts and sensor noise. Therefore, the Manipulator itself does not explicitly separate motion from noise, and the amplified features still require subsequent global-context refinement.
\vspace{-0.2cm}

\subsection{Mamba-based Global Consistency}
\noindent \textbf{Global Context Modeling.} Indiscriminate amplification often exacerbates imaging noise. To address this, the Mamba-based Sequence Awareness (MMSA) encoder processes the manipulated feature $\tilde{F}$ using $N_1 = 12$ Vision Mamba blocks. The core 2D Selective Scan mechanism functions as a spatially-variant filtering process that leverages the input-dependent selectivity term $\Delta$ to favor coherent motion patterns over stochastic noise. Since geometric motion exhibits spatial coherence while noise manifests as stochastic fluctuations, this mechanism enables the hidden states to preferentially retain motion-relevant structures while reducing random artifacts. Consequently, MMSA refines the amplified latent motion into a globally consistent representation with linear complexity, yielding the refined feature $Z$, while static high-frequency details are mainly preserved by the parallel SFR branch.

\subsection{Static Feature Refinement and Fusion}
\noindent \textbf{Static Feature Refiner (SFR).} Deep encoders tend to lose high-frequency texture information. To mitigate this information loss, a parallel refinement branch is employed that operates directly on the shallow, high-resolution features $F_a^h$ and $F_b^h$. The SFR module, represented by the function $\mathcal{S}$, extracts texture residuals to stabilize the static background. As depicted in the architecture, the module employs a subtraction mechanism to refine the source features. The refined static feature $F_s^r$ is computed as:
\begin{equation}
    F_s^r = F_a^h - \mathcal{S}(F_a^h - F_b^h)
\end{equation}
This subtractive refinement stabilizes the background and suppresses artifacts under large magnification factors. Unlike a direct skip connection, SFR preserves static high-frequency details while reducing the direct reintroduction of unmagnified motion and sensor noise; Section~V-D verifies its necessity within our framework.

\noindent \textbf{Fusion and Reconstruction.} To generate the final video frame, the globally consistent motion features from the primary stream (upsampled to match resolution, denoted as $Z^u$) are combined with the refined static details. A fusion decoder $\mathcal{D}$ integrates these components to reconstruct the output $\hat{I}_a$:
\begin{equation}
    \hat{I}_a = \mathcal{D}(Z^u + F_s^r)
\end{equation}
By explicitly combining global motion coherence with local textural fidelity, our approach supports high-fidelity magnification.

\section{Experiments}

\subsection{Experiment Setup}
\noindent \textbf{Real-World Datasets.} Following the standard evaluation protocol established by LBVMM~\cite{oh2018learning}, two widely adopted benchmarks are utilized: (a) the Static dataset and (b) the Dynamic dataset, which contain 10 and 6 classic videos respectively. These sequences cover scenarios ranging from subtle breathing motions to large camera movements. Additionally, a self-collected Industrial Rotor dataset is introduced, comprising 6 videos of mechanical rotors operating at varying speeds to test high-frequency vibration analysis. 

\noindent \textbf{Synthetic Datasets.} Four custom datasets were constructed (10k samples each). Synthetic-I contains translational motion, while Synthetic-II introduces rigid body rotation. Synthetic-III and Synthetic-IV further incorporate Poisson noise and Gaussian blur. On these datasets, models are evaluated by amplifying inputs with specific amplification factors ($\alpha$) against ground truth. 

\noindent \textbf{Implementation Details.} Our method is implemented in PyTorch and trained end-to-end under two training-data settings: the standard LBVMM-style synthetic protocol~\cite{oh2018learning}, reported as ``Ours (Original Data)'' in Table~I, and our proposed Geo-200K dataset, reported as ``Ours (Geo-200K)''. The Adam optimizer is utilized with an initial learning rate of $1\times10^{-4}$. The objective function combines $\mathcal{L}_1$ reconstruction loss, LPIPS perceptual loss, and regularization. Detailed settings are provided in the Supplementary Material.

\noindent \textbf{Evaluation Metrics.} For synthetic datasets, we employ RMSE, PSNR, and LPIPS~\cite{lpipszhang2018unreasonableeffectivenessdeepfeatures} to evaluate reconstruction accuracy and perceptual fidelity. For real-world datasets, full-reference quantitative evaluation is not applicable because magnified ground-truth targets are unavailable. Therefore, we report qualitative results by examining spatial regions and spatiotemporal (ST) slices. User-study details are provided in the Supplementary Material.

\begin{figure*}[!htbp]
  \centering
  \includegraphics[width=0.9\linewidth]{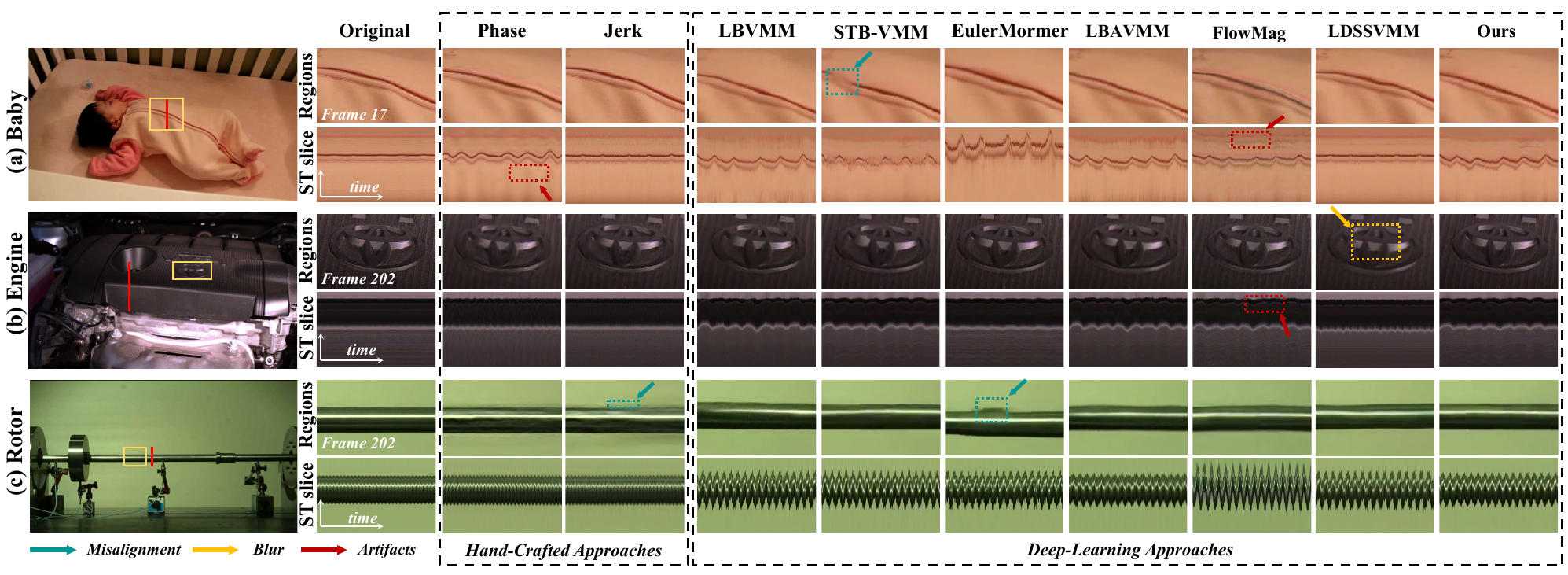}
  \vspace{-0.4cm}
\caption{Qualitative comparison on real-world videos: (a) Baby ($\alpha=20$), (b) Engine ($\alpha=10$), and (c) Rotor ($\alpha=10$). Compared with prior methods, GeoMag better preserves structural continuity while reducing blur and artifacts. More examples are provided in the Supplementary Material.}
   \vspace{-0.45cm}
\label{fig:Main_Compare}
\end{figure*}

\subsection{Quantitative Comparisons}
\noindent \textbf{Quantitative Evaluation.} We evaluate the proposed framework on four custom synthetic datasets (Synthetic-I to IV) totaling 40,000 samples. Table~\ref{tab:MainCompTable} disentangles the contributions of the dataset and the model design; ``Ours (Original Data)'' and ``Ours (Geo-200K)'' denote the same architecture trained on the standard LBVMM-style protocol~\cite{oh2018learning} and Geo-200K, respectively. For fairness, only baselines with publicly available and protocol-consistent training pipelines are retrained on Geo-200K, while the remaining methods are evaluated using their official implementations. We focus on directly comparable RGB-based VMM methods. Approaches relying on different sensing modalities (e.g., event-based inputs) or lacking reproducible public weights are excluded from the main comparison to avoid introducing confounding factors.

Comparing Group I and Group II shows the effect of our data pipeline. Retraining baselines such as LBVMM and STB-VMM on Geo-200K yields consistent gains across metrics; notably, STB-VMM achieves a $+0.68$ dB PSNR improvement on Synthetic-I, suggesting that Geo-200K provides richer kinematic supervision than standard protocols.

Group III further demonstrates the strong performance of GeoMag. A clear performance gap is observed on Synthetic-II (Rotation). While existing baselines show limitations in maintaining structural rigidity under non-linear rotation, GeoMag achieves the lowest LPIPS (0.16) and RMSE. This suggests that the global context modeling in GeoMag is more effective for preserving rotational trajectories than the window-based attention or local convolutions used in prior methods. GeoMag also shows strong robustness under Poisson noise (Synthetic-III) and Gaussian blur (Synthetic-IV), maintaining high fidelity (PSNR $>$ 21.3 dB) while competitive baselines such as EulerMormer exhibit noticeable performance drops.

\noindent \textbf{Efficiency Analysis.} Table~\ref{tab:efficiency_time} compares computational efficiency. While CNNs are fast but often lack global structural consistency, GeoMag achieves a favorable trade-off. Capitalizing on linear complexity, it requires $\sim5\times$ fewer GFLOPs than STB-VMM (296.1G vs. 1584G). Consequently, GeoMag processes a 300-frame 720p video in just 13.9s, significantly faster than STB-VMM (31.6s) and EulerMormer (50.0s).

\subsection{Qualitative Analysis}
\noindent \textbf{Magnification visualization comparisons.}
Fig.~\ref{fig:Main_Compare} presents visual comparisons across real-world scenarios. Hand-crafted methods (Phase, Jerk) fail to balance performance, yielding either imperceptible motion or excessive ringing artifacts. Conventional CNN-based approaches (e.g., LDSSVMM) tend to be conservative, suffering from texture blurring as indicated by the yellow box in the \textit{Engine} sequence. Furthermore, distinct visual artifacts appear in flow-based methods; as shown by the red dotted box in the Baby case, FlowMag misinterprets texture as motion, generating unnatural ripples. Notably, advanced baselines (STB-VMM, EulerMormer) exhibit severe geometric inconsistencies; as highlighted by the green arrows in the Rotor sequence, these methods introduce spatial discontinuities where rigid structures break. In contrast, GeoMag produces more reliable magnification results, reducing artifacts while preserving better structural integrity. Evident in the Rotor spatial regions, GeoMag maintains a continuous trajectory, contrasting with the fractured baselines. Additional visual comparisons and user-study results are provided in the Supplementary Material.
\vspace{-0.5cm}
\begin{table}[!htbp]
\centering
\caption{Efficiency comparison in GFLOPs, FPS, and processing time for a 300-frame video at $384\times384$ and $720\times720$.}
\label{tab:efficiency_time}
\vspace{-0.2cm}
\setlength{\tabcolsep}{2pt} 
\resizebox{\linewidth}{!}{
\begin{tabular}{l c cc cc cc} 
\toprule
\multirow{2}{*}{\textbf{Method}} & \multirow{2}{*}{\textbf{Params (M)}} & \multicolumn{2}{c}{\textbf{GFLOPs} $\downarrow$} & \multicolumn{2}{c}{\textbf{FPS} $\uparrow$} & \multicolumn{2}{c}{\textbf{Time (s)} $\downarrow$} \\
\cmidrule(lr){3-4} \cmidrule(lr){5-6} \cmidrule(lr){7-8}
 & & \textbf{384p} & \textbf{720p} & \textbf{384p} & \textbf{720p} & \textbf{384p} & \textbf{720p} \\
\midrule
\multicolumn{8}{c}{\textit{CNN-Based Models}} \\
\midrule
LBVMM       & 0.97 & 55.2 & 194.2 & 137.3 & 42.7 & 2.2 & 7.0 \\
LBAVMM      & 0.97 & 55.2 & 194.2 & 136.1 & 43.1 & 2.2 & 7.0 \\
LDSSVMM       & 0.09 & 13.89 & 48.81  & 92 & 26 & 3.26 & 11.54 \\
\midrule
\multicolumn{8}{c}{\textit{Flow-Based Models}} \\
\midrule
FlowMag     & 17.29 & 93.4 & 328.4 & 227.0 & 58.6 & 1.3 & 5.1 \\
\midrule
\multicolumn{8}{c}{\textit{Attention-Based Models}} \\
\midrule
STB-VMM     & 31.24 & 445.4 & 1584  & 28.3 & 9.5  & 10.6 & 31.6 \\
EulerMormer & 1.51  & 61.5  & 216.2 & 22.2 & 6.0  & 13.5 & 50.0 \\
\midrule
\rowcolor{gray!15}
\textbf{Ours} & \textbf{30.73} & \textbf{84.2 }& \textbf{296.1} & \textbf{80.9} & \textbf{21.6} & \textbf{3.7} & \textbf{13.9} \\
\bottomrule
\end{tabular}
}
\end{table}

\subsection{Ablation Studies}
\noindent \textbf{Effect of Loss Function.} Table \ref{tab:ablation} summarizes the ablation study on loss terms. The baseline using only $\mathcal{L}_{1}$ shows limited performance. Integrating $\mathcal{L}_{reg}$ improves pixel-wise accuracy (RMSE and PSNR), while incorporating $\mathcal{L}_{lpips}$ enhances perceptual quality (LPIPS). The full model, which combines all terms, achieves the best overall performance across all metrics (RMSE 24.22, PSNR 21.09, LPIPS 0.18), demonstrating the complementarity and importance of the proposed loss terms for high-fidelity generation.

\begin{table}[!htbp]
\centering
\renewcommand{\arraystretch}{1}
\setlength{\tabcolsep}{8pt}

\vspace{-0.25cm}
\caption{Quantitative evaluation of the effectiveness of different loss terms}
\vspace{-0.25cm}
\begin{tabular}{ccc|ccc}
\toprule
\multicolumn{3}{c|}{Loss Components} & \multicolumn{3}{c}{Metrics} \\ 
\cmidrule(lr){1-3} \cmidrule(lr){4-6}
$\mathcal{L}_{1}$ & $\mathcal{L}_{reg}$ & $\mathcal{L}_{lpips}$ & RMSE $\downarrow$ & PSNR $\uparrow$ & LPIPS $\downarrow$ \\ 
\midrule
\checkmark & & & 29.96 & 19.01 & 0.36 \\
\checkmark & \checkmark & &27.38 & 19.79 & 0.30  \\
\checkmark & & \checkmark &  28.46 &  19.46&    0.23\\
\checkmark & \checkmark & \checkmark & \textbf{24.22} & \textbf{21.09} & \textbf{0.18} \\ 
\bottomrule
\end{tabular}
\label{tab:ablation}
\end{table}

\noindent \textbf{Validation of DFE and MMSA Encoders.} To validate the efficacy of the proposed encoding modules, we conduct an ablation study by replacing both the Deep Feature Extractor (DFE) and the MMSA Encoder with Transformer and CNN variants under the same overall framework. As presented in Table \ref{tab:ablation2}, GeoMag outperforms both baselines (e.g., achieving $0.18$ LPIPS compared to $0.34$ for the Transformer variant). This result supports the effectiveness of the proposed encoders compared with conventional backbone alternatives. Detailed configurations are provided in the Supplementary Material.

\noindent \textbf{Impact of the SFR.} As shown in Table \ref{tab:ablation2}, removing the Static Feature Refiner ($\mathcal{S}$) causes marked degradation, increasing RMSE from $24.22$ to $29.42$. This result highlights the role of SFR in stabilizing static regions and preserving high-frequency details that are important for perceptual quality.
\vspace{-0.35cm}
\begin{table}[!htbp]
  \centering
 \caption{Ablation study on architectural components.
The effectiveness of GeoMag backbone is validated by comparing it with CNN and Transformer counterparts, and the contribution of SFR is verified.}
  \vspace{-0.25cm}
  \label{tab:ablation2}
  \renewcommand{\arraystretch}{1.1}
  \setlength{\tabcolsep}{3.0mm}
  \begin{tabular}{l|ccc}
    \toprule
    \textbf{Method / Variant} & \textbf{RMSE} $\downarrow$ & \textbf{PSNR} $\uparrow$ & \textbf{LPIPS} $\downarrow$ \\

    \midrule
      CNN Variant  & 32.72    & 17.39             &   0.29              \\
     Transformer Variant   & 29.60          & 18.61          & 0.34           \\
         \midrule
   GeoMag (w/o SFR)                  & 29.42          & 18.85          & 0.26           \\
    \midrule
    \textbf{GeoMag}    & \textbf{24.22} & \textbf{21.09} & \textbf{0.18}  \\
    \bottomrule
  \end{tabular}
\end{table}

\vspace{-0.3cm}
\section{Conclusion}
\vspace{-0.05cm}
This paper presents GeoMag, a framework that effectively mitigates the trade-off between long-range dependency modeling and computational costs. By incorporating State Space Models with selective scanning, global geometric consistency is captured while maintaining linear complexity. To address current data limitations, the Geo-200K dataset was established, explicitly constructed to simulate complex geometric transformations. Extensive experiments demonstrate that our approach outperforms state-of-the-art methods, achieving superior visual fidelity with approximately 5$\times$ fewer GFLOPs than Transformer-based baselines. Ultimately, this work provides a strong benchmark and a high-quality data foundation for efficient motion magnification research.

\bibliographystyle{IEEEbib}
\bibliography{icme2026references}

@article{wu2012eulerianfluid,
  title={Eulerian video magnification for revealing subtle changes in the world},
  author={Wu, Hao-Yu and Rubinstein, Michael and Shih, Eugene and others},
  journal={TOG},
  volume={31},
  number={4},
  year={2012}
}

@article{STBVMM_LADOROIGE2023110493,
  title={STB-VMM: Swin Transformer based Video Motion Magnification},
  author={Lado-Roigé, Ricard and Pérez, Marco A.},
  journal={KBS},
  year={2023}
}

@inproceedings{flowmag_2023,
  title={Self-Supervised Motion Magnification by Backpropagating Through Optical Flow},
  author={Pan, Zhaoying and Geng, Daniel and Owens, Andrew},
  booktitle={NeurIPS},
  year={2023}
}

@inproceedings{bai2021micro,
  title={Micro-expression recognition based on video motion magnification and pre-trained neural network},
  author={Bai, Mengjiong and Goecke, Roland and Herath, Damith},
  booktitle={ICIP},
  year={2021}
}

@inproceedings{gu2024mamba,
  title={Mamba: Linear-time sequence modeling with selective state spaces},
  author={Gu, Albert and Dao, Tri},
  booktitle={COLM},
  year={2024}
}

@inproceedings{zhou2022audio,
  title={Audio--visual segmentation},
  author={Zhou, Jinxing and Wang, Jianyuan and Zhang, Jiayi and others},
  booktitle={ECCV},
  year={2022}
}

@article{zhang2023hybrid,
  title={Hybrid Eulerian--Lagrangian framework for structural full-field vibration quantification and modal shape visualization},
  author={Zhang, Dashan and Zhu, Andong and Gong, Xinlong and others},
  journal={Measurement},
  volume={219},
  year={2023}
}

@article{shabi2020motion,
  title={Motion magnification analysis of microscopy videos of biological cells},
  author={Shabi, Oren and Natan, Sari and Kolel, Avraham and others},
  journal={PLOS One},
  volume={15},
  number={11},
  year={2020}
}

@article{liu2005motion,
  title={Motion magnification},
  author={Liu, Ce and Torralba, Antonio and Freeman, William T. and others},
  journal={TOG},
  volume={24},
  number={3},
  year={2005}
}

@inproceedings{wadhwa2014riesz,
  title={Riesz pyramids for fast phase-based video magnification},
  author={Wadhwa, Neal and Rubinstein, Michael and Durand, Fr{\'e}do and others},
  booktitle={ICCP},
  year={2014}
}

@article{le2019seeing,
  title={Seeing the invisible: Survey of video motion magnification and small motion analysis},
  author={Le Ngo, Anh Cat and Phan, Rapha{\"e}l C-W},
  journal={CSUR},
  volume={52},
  number={6},
  year={2019}
}

@inproceedings{oh2018learning,
  title={Learning-based video motion magnification},
  author={Oh, Tae-Hyun and Jaroensri, Ronnachai and Kim, Changil and others},
  booktitle={ECCV},
  year={2018}
}

@inproceedings{deng2024coconut,
  title={Coconut: Modernizing coco segmentation},
  author={Deng, Xueqing and Yu, Qihang and Wang, Peng and others},
  booktitle={CVPR},
  year={2024}
}

@article{openiamges2020open,
  title={The open images dataset v4: Unified image classification, object detection, and visual relationship detection at scale},
  author={Kuznetsova, Alina and Rom, Hassan and Alldrin, Neil and others},
  journal={IJCV},
  volume={128},
  number={7},
  year={2020}
}

@inproceedings{wang2024eulermormer,
  title={Eulermormer: Robust eulerian motion magnification via dynamic filtering within transformer},
  author={Wang, Fei and Guo, Dan and Li, Kun and others},
  booktitle={AAAI},
  year={2024}
}

@inproceedings{2018jerk,
  title={Jerk-aware video acceleration magnification},
  author={Takeda, Shoichiro and Okami, Kazuki and Mikami, Dan and others},
  booktitle={CVPR},
  year={2018}
}

@inproceedings{2024Axial,
  title={Learning-based Axial Video Motion Magnification},
  author={Kwon, Byung-Ki and Oh, Hyun-Bin and Kim, Jun-Seong and others},
  booktitle={ECCV},
  year={2024}
}

@inproceedings{wang2024frequency,
  title={Frequency decoupling for motion magnification via multi-level isomorphic architecture},
  author={Wang, Fei and Guo, Dan and Li, Kun and others},
  booktitle={CVPR},
  year={2024}
}

@inproceedings{singh2023multi,
  title={Multi domain learning for motion magnification},
  author={Singh, Jasdeep and Murala, Subrahmanyam and Kosuru, G.},
  booktitle={CVPR},
  year={2023}
}

@article{wadhwa2013phase,
  title={Phase-based video motion processing},
  author={Wadhwa, Neal and Rubinstein, Michael and Durand, Fr{\'e}do and others},
  journal={TOG},
  volume={32},
  number={4},
  year={2013}
}

@inproceedings{fang2023rmesICME,
  title={RMES: real-time micro-expression spotting using phase from Riesz pyramid},
  author={Fang, Yini and Deng, Didan and Wu, Liang and others},
  booktitle={ICME},
  year={2023}
}

@inproceedings{zhu2024visionmambaefficientvisual,
  title={Vision Mamba: Efficient Visual Representation Learning with Bidirectional State Space Model},
  author={Zhu, Lianghui and Liao, Bencheng and Zhang, Qian and others},
  booktitle={ICML},
  year={2024}
}

@inproceedings{Xiaoicmemicro,
  title={ESTME: Event-driven Spatio-temporal Motion Enhancement for Micro-Expression Recognition},
  author={Xiao, Peilin and Zhang, Yueyi and Kai, Dachun and others},
  booktitle={ICME},
  year={2024}
}

@inproceedings{guo2024mambairsimplebaselineimage,
  title={MambaIR: A simple baseline for image restoration with state-space model},
  author={Guo, Hang and Li, Jinmin and Dai, Tao and others},
  booktitle={ECCV},
  year={2024}
}

@inproceedings{lpipszhang2018unreasonableeffectivenessdeepfeatures,
  title={The Unreasonable Effectiveness of Deep Features as a Perceptual Metric},
  author={Zhang, Richard and Isola, Phillip and Efros, Alexei A. and others},
  booktitle={CVPR},
  year={2018}
}

@misc{2505.17476,
  title={The Coherence Trap: When MLLM-Crafted Narratives Exploit Manipulated Visual Contexts},
  author={Zhang, Yuchen and Wang, Yaxiong and Wu, Yujiao and others},
  note={arXiv},
  year={2025}
}

@misc{2603.01993,
  title={Process Over Outcome: Cultivating Forensic Reasoning for Generalizable Multimodal Manipulation Detection},
  author={Zhang, Yuchen and Wang, Yaxiong and Han, Kecheng and others},
  note={arXiv},
  year={2026}
}

@article{LDSSVMM2025,
  title={Learnable Directional Scale Space Filters for Video Motion Magnification},
  author={Singh, Jasdeep and Kumar Vipparthi, Santosh and Murala, Subrahmanyam and others},
  journal={KBS},
  volume={330},
  year={2025}
}

@inproceedings{icmeMicro,
  title={TGMAE: Self-supervised Micro-Expression Recognition with Temporal Gaussian Masked Autoencoder},
  author={Liu, Shifeng and Mao, Xinglong and Zhao, Sirui and others},
  booktitle={ICME},
  year={2024}
}

@inproceedings{icmeGraph,
  title={Adaptive Graph Attention Network with Temporal Fusion for Micro-Expressions Recognition},
  author={Zhang, Yiming and Wang, Hao and Xu, Yifan and others},
  booktitle={ICME},
  year={2023}
}

\end{document}